%% file: root.tex
\documentclass[letterpaper, 10 pt, conference]{ieeeconf}  

\IEEEoverridecommandlockouts                              

\usepackage[utf8]{inputenc} 
\usepackage[T1]{fontenc}    
\usepackage{hyperref}       
\usepackage{url}            
\usepackage{booktabs}       
\usepackage{amsfonts}       
\usepackage{nicefrac}       
\usepackage{microtype}      
\usepackage{lipsum}
\usepackage{comment}
\usepackage{textcomp}
\usepackage{xcolor}
\usepackage{graphicx}
\usepackage{color,soul}
\usepackage{amsmath}
\usepackage{subcaption}
\usepackage{siunitx}
\newcommand{\norm}[1]{\left\lVert#1\right\rVert}
\title{\LARGE \bf
Low Dimensional State Representation Learning with Robotics Priors in Continuous Action Spaces
}

\author{Nicolò Botteghi$^{1}$, Khaled Alaa$^{2}$, Mannes Poel$^{1}$, Beril Sirmacek$^{3}$, Christoph Brune$^{1}$, \\ Abeje Mersha$^{4}$, and Stefano Stramigioli$^{1}$
\thanks{$^{1}$Nicolò Botteghi is with Faculty of Electrical Engineering, Mathematics and Computer Science,
        University of Twente, Enschede, The Netherlands
        {\tt\small n.botteghi@utwente.nl}}%
\thanks{$^2$ Khaled Alaa is with Intelligent Driving Functions R\&D Center, IAV GmbH (Volkswagen Group), Berlin, Germany
{\tt\small khaled.mustafa@iav.de}}
\thanks{$^{3}$Beril Sirmacek is with the Department of Computer Science, J\"{o}nk\"{o}ping University, J\"{o}nk\"{o}ping, Sweden}
\thanks{$^4$ Abeje Mersha is with the Research Group of Mechatronics, Saxion University of Applied Sciences, Enschede, The Netherlands.}
}

\begin{document}

\maketitle
\thispagestyle{empty}
\pagestyle{empty}

\begin{abstract}


Autonomous robots require high degrees of cognitive and motoric intelligence to come into our everyday life. In non-structured environments and in the presence of uncertainties, such degrees of intelligence are not easy to obtain. Reinforcement learning algorithms have proven to be capable of solving complicated robotics tasks in an end-to-end fashion without any need for hand-crafted features or policies.  
Especially in the context of robotics, in which the cost of real-world data is usually extremely high, reinforcement learning solutions achieving high sample efficiency are needed. In this paper, we propose a framework combining the learning of a low-dimensional state representation, from high-dimensional observations coming from the robot's raw sensory readings, with the learning of the optimal policy, given the learned state representation.  
We evaluate our framework in the context of mobile robot navigation in the case of continuous state and action spaces. Moreover, we study the problem of transferring what learned in the simulated virtual environment to the real robot without further retraining using real-world data in the presence of visual and depth distractors, such as lighting changes and moving obstacles.  A video of our experiments can be found at: \url{https://youtu.be/rUdGPKr2Wuo}.

\end{abstract}

\section{INTRODUCTION}\label{sec:introduction}
\input{introduction}

\section{BACKGROUND}\label{sec:background}
\input{background}

\section{RELATED WORK}\label{sec:related_work}
\input{related_work}

\section{METHODOLOGY}\label{sec:methodology}
\input{methodology}

\section{EXPERIMENTAL DESIGN}\label{sec:experimental_design}
\input{experimental_design}

\section{RESULTS AND DISCUSSIONS}\label{sec:results}
\input{results}

\section{CONCLUSIONS}\label{sec:conclusion}
\input{conclusion}

\end{document}

%% file: introduction.tex
Robots are often envisioned as replacements for humans for the most tedious or dangerous jobs.  Robotics research has tremendously progressed in the last decades. Robots can now be programmed for solving different tasks, from everyday life simplest jobs, e.g. vacuum cleaning, or grass-cutting, to complex industrial applications, e.g. car assembly, smart warehouse, or inspection of plants. However, many steps have yet to be taken to achieve high degrees of cognitive and motoric intelligence. These automated solutions require a vast amount of prior knowledge of the tasks, and they are often brittle in all the scenarios in which the robots have an imperfect and limited perception, inaccurate models of the world and uncertainties in the motion. 

Reinforcement learning \cite{Sutton1998}, or RL, is the machine learning field studying the problem of optimal sequential decision making in the presence of uncertainties through the \textit{trial-and-error} paradigm. Reinforcement learning has shown to be able to learn complex behaviours directly from high-dimensional input data, e.g. raw pixel inputs, in different domains such as robotics \cite{e2c}, \cite{e2evisuo} and video-games \cite{DeepmindAtari2013}, \cite{AlphaGo}, \cite{AlphaZero}. Thus, reinforcement learning has the potential to achieve high degrees of motoric and cognitive intelligence. 

\begin{figure*}[ht!]
    \centering
    \includegraphics[page=1,width=0.8\linewidth]{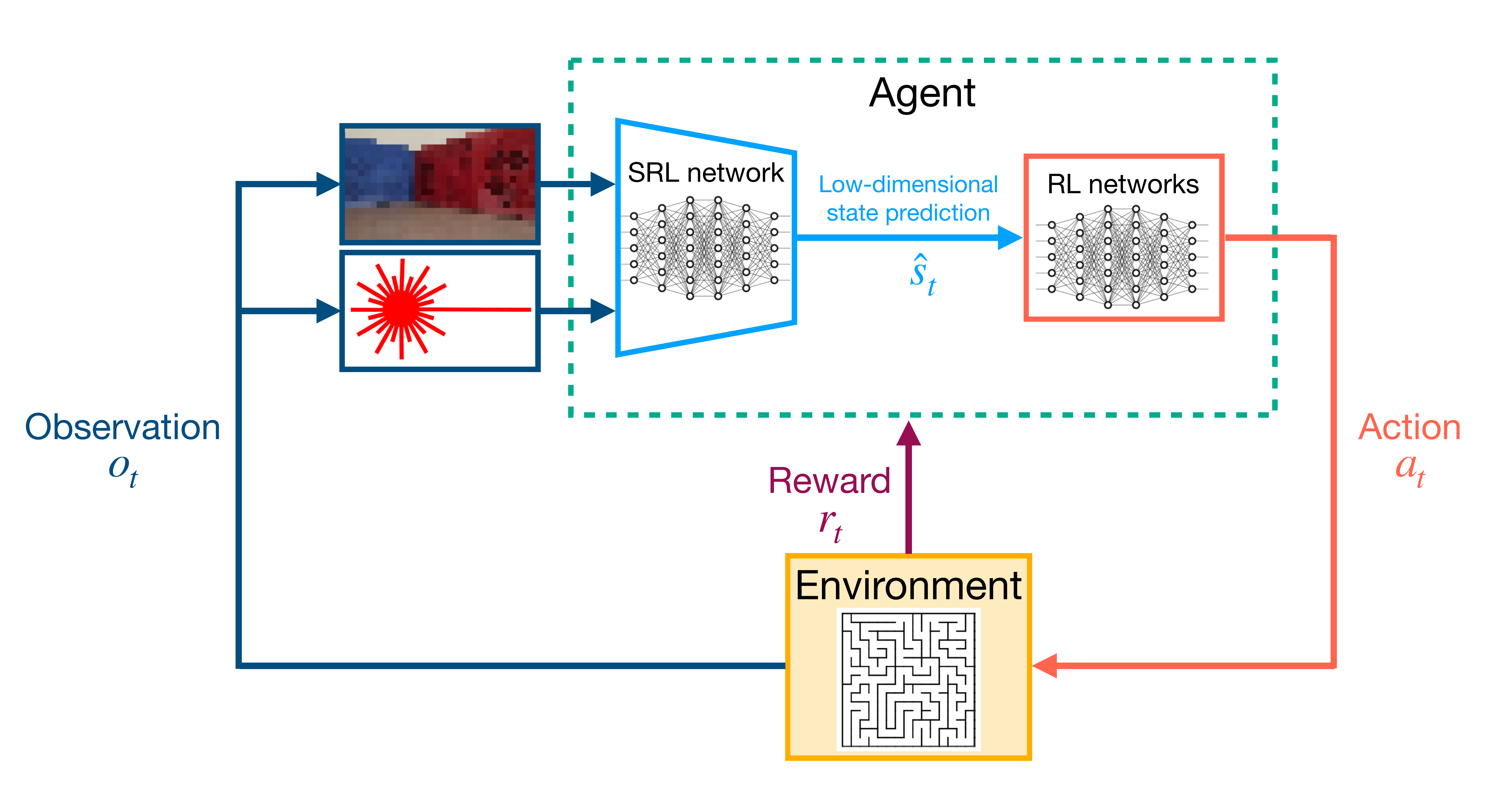}
    \caption{Proposed framework combining state representation learning and reinforcement learning.}
    \label{fig:state_representation_framework}
\end{figure*}

Learning from high-dimensional inputs, or observations, comes at the price of low sample efficiency and high computational load. 
The problem of high sample-efficiency of the algorithms is of crucial importance in robotics due to the cost of real hardware and simulators \cite{RoboticsRLsurvey}, \cite{RoboticsRLApplications}. 
Unfortunately, simulators rely on mathematical models (e.g. physics) that approximate the real world. These approximations make transfer learning a big challenge in many cases. This problem is referred to, in literature, as the \textit{simulation-to-reality gap} \cite{RoboticsRLsurvey}.
Moreover, when learning in virtual simulated environments, we have access to the true state of the environment. For example, in a mobile robot navigation task, this corresponds to the knowledge of the true pose of the robot and the location of the navigation target. However, when aiming to solve real-world challenges, we must rely on raw sensory data, such as RGB cameras or laser range (LiDAR) sensors. Sensory data are often noisy, partial and high-dimensional, and therefore it is difficult to deal with them. 

The state space can be manually hand-crafted to contain only the relevant features and consequently reduce its dimensionality to overcome the issue of learning policies directly from raw observations and improve sample efficiency. While this step is possible for simple tasks, designing the state space is a non-trivial challenge in real-world problems. Moreover, hand-crafted features are usually brittle to changes in environments such as lighting conditions or non-modelled dynamics. The robustness of feature extraction methods is especially critical in robotics reinforcement learning that heavily relies on virtual simulators to reduce the cost of real-world samples. Therefore, we need methods for extracting task-relevant features in robust and sample-efficient ways.

State representation learning, or SRL, \cite{srlsurvey} is the name given to all the methods learning to extract informative and compact state representations from high-dimensional observations to facilitate the learning of the policy of the reinforcement learning algorithms. Instead of directly solving the problem of learning the optimal policy, mapping observations to actions, it is beneficial, for sample-efficiency and robustness of the learned policy, to first learn a low-dimensional representation of the observations, i.e. the state representation, then learn the optimal policy directly from such representation. The mapping observations-to-states can be learned through supervised or unsupervised learning. In the former case, true state values are needed. Unfortunately, these may not always be available; therefore, the focus is on unsupervised learning methods.

Robots live in spaces governed by physical laws. These spaces have a lower dimensionality than the space of raw sensory observations. We can exploit such knowledge and represent the sensory information into learned state spaces of low-dimensionality.
Moreover, additional and general \textit{priors} \cite{bengio2013}, i.e. prior knowledge, of the world can be used to aid the learning of meaningful state representations, such as smoothness of the state changes or the relation between action magnitude and state changes. These priors can be used to shape the loss functions used for learning such mappings. In the particular case of robotics, the authors of \cite{Jonschkowski2015} propose the so-called \textit{robotics priors}, i.e. loss functions that are shaped to incorporate physical knowledge of the world and used to constrain and steer the learning of the state representation loosely. The robotics priors are means to learn meaningful state representations in all the cases in which physical laws govern the environment, and they consequently allow robustness and sample efficient learning of the reinforcement learning policy.

This research proposes a framework for state representation learning and reinforcement learning to efficiently solve mobile robots navigation tasks directly from raw sensory information. This framework is shown in Figure \ref{fig:state_representation_framework}. Our research contributes to improving the sample efficiency of the reinforcement learning algorithm, increasing robustness and interpretability of the learned state features, combining the use of deep learning with prior knowledge, and bridging the simulation-to-reality gap.  
In particular:
\begin{itemize}
    \item We study the problem of unsupervised learning, i.e. without the need for expensive labelled data, of meaningful and interpretable state representations using the robotics priors in continuous state and action spaces. Continuous action spaces allow smoother and more advanced trajectories, and therefore should be preferred for robotics.
    \item We introduce a new set of robotics priors, i.e. loss functions for learning the state representation, that exploit the underlying structure of continuous action spaces for obtaining smoother and more informative state representations. The new loss functions do not only contain information of states but also actions.
    \item Eventually, we show that the proposed framework allows the transfer of the policies and representations learned in virtual simulation environments to the real robot without further retraining using real-world data.
\end{itemize}

The rest of the paper is organized as follows: Section \ref{sec:background} presents the background information to reinforcement learning and state representation learning and Section \ref{sec:related_work} discusses the related work to this research. Then, Section \ref{sec:methodology} explains the methodology and the proposed approach, followed by Section \ref{sec:experimental_design} showing the experimental design. Section \ref{sec:results} presents the results and discusses them. Eventually, Section \ref{sec:conclusion} concludes the paper.

%% file: background.tex
\subsection{Markov Decision Processes}

Markov Decision Processes, or MDPs, \cite{mdp} are a mathematical framework for studying and solving sequential decision making processes. In any MDP, we can identify two major entities: the agent and the environment. The agent, or the decision-maker, tries to learn the optimal way of behaving, i.e. the optimal policy, while the environment corresponds to the world in which the agent lives. Formally, an MDP $\mathcal{M}$ is a tuple $\left(\mathcal{S}, \mathcal{A}, \text{T}, \text{R} \right)$, where $\mathcal{S}$ is the set of states, $\mathcal{A}$ is the set of actions, $\text{T}(s,a): \mathcal{S} \times \mathcal{A} \longrightarrow [0,1]$ is the transition function determining the evolution of the states and $\text{R}(s,a): \mathcal{S} \times \mathcal{A} \longrightarrow \mathbb{R}$ is the reward function evaluating the quality of the actions taken by the agent. 
When the dynamic model of the MDP,  i.e. the transition function $\text{T}$ and the reward function $\text{R}$, is not known a priori, the agent has to discover the best acting strategy by interacting with the environment through the trial-and-error paradigm.

\subsection{Reinforcement Learning}
Reinforcement learning, or RL, \cite{Sutton1998} is the name given to the collection of algorithms solving sequential decision-making processes without any knowledge of the underlying MDP. The aim of any reinforcement learning algorithm is finding the optimal policy $\pi$, mapping states to actions, for maximizing the total cumulative return $R_t$ in Equation (\ref{cumulative_rew}).
\begin{equation}
R_t = \Sigma_{t=0}^{\infty} \gamma^{t} r_{t+1}
\label{cumulative_rew}
\end{equation}
where the subscript $t$ indicates the time steps, $\gamma$ is the discount factor, and $r_t$ is the instantaneous reward received by the agent at time step $t$.

\subsubsection*{Deep Deterministic Policy Gradient}

Deep Deterministic Policy Gradient, or DDPG, \cite{DDPG2015} is an actor-critic reinforcement learning algorithm that extends Deep Q-Network, or DQN, \cite{DeepmindAtari2013} to continuous action spaces. The actor, i.e. the policy $\pi$, chooses an action for each input state, while the critic, i.e. the action-value function $\text{Q}$, evaluates the performance of the actor. The actor and the critic are modelled using two neural networks, respectively $\theta^{\pi}$ and $\theta^\text{Q}$. To improve the training stability, DDPG uses a copy of the critic and the actor networks, parametrized respectively by $\theta^{\text{Q}'}$ and $\theta^{\pi'}$, that are updated with a slower frequency than the actor and the critic accordingly to Equation (\ref{target_nns_update}).
\begin{equation}
\begin{split}
\theta^{\text{Q}'} &= \rho \theta^{\text{Q}} + (1-\rho)\theta^{\text{Q}'}\\
\theta^{\pi'} &= \rho \theta^{\pi} + (1-\rho)\theta^{\pi'} \\
\end{split}
\label{target_nns_update}
\end{equation}
where $\rho$ is a constant determining the speed of the updates of the target networks.
To improve the policy, first, the parameters of the critic $\theta^\text{Q}$, are adjusted according to the mean square error loss between the predicted Q-value $\text{Q}(s_t,a_t|\theta_i^\text{Q})$ and the target Q-value $y_t=r(s_t,a_t)+\gamma \text{Q}(s_{t+1},a_{t+1}|\theta_i^{\text{Q}'})$ estimated via the target network, as shown in Equation (\ref{loss}).
\begin{equation}
   \mathcal{L}(\theta^\text{Q}) = \mathbb{E}_{\pi}[(\text{Q}(s_t,a_t|\theta^\text{Q})-y_t)^2]
   \label{loss}
\end{equation}
The estimate of the state-action value function \text{Q} is used to update the parameters of the actor network, $\theta^\pi$, that are adjusted in the direction of the gradient of the expected return  $\nabla_{\theta^\pi}J(\pi_\theta)$, shown in Equation (\ref{return_gradient}).
\begin{equation}
    \nabla_{\theta^\pi}J(\pi_\theta)=\mathbb{E}_{\pi}[\nabla_a \text{Q}(s_t,\pi(s_t|\theta^\pi)|\theta^\text{Q})\nabla_{\theta^\pi}\pi(s_t|\theta^\pi)]
    \label{return_gradient}
\end{equation}
In this way, the gradient of the critic guides the improvements of the actor.

\subsection{State Representation Learning}

In many interesting applications, the state of the environment is not directly observable by the reinforcement learning agent. However, the agent can only perceive the environment through high-dimensional observations, such as sensory reading, e.g. RGB images or LiDAR data in the case of a mobile robot. While reinforcement learning algorithms can learn the relevant information from the inputs using only the reward signal, it is possible and convenient, for the sake of sample-efficiency, generalization and robustness \cite{srlsurvey}, to first learn a compact and low-dimensional state representation encoding all the relevant information for solving the given task and then learn the optimal policy directly using the learned state representation.  

State representation learning methods aim at learning a meaningful state representation from high-dimensional observations. Here, we focus on all the methods employing unsupervised or self-supervised learning approaches for obtaining such representations, where the mapping is learned through neural networks.
To learn the observation-state mapping $\phi$, the complete history of observations $o_{1:t}$, the actions taken $a_{1:t}$ and the rewards received $r_{1:t}$ from the environment can be used, as shown in Equation (\ref{eq:form}). 
\begin{equation}
\hat{s}_t = \phi (o_{1:t},a_{1:t},r_{1:t})
\label{eq:form}
\end{equation}
where $\hat{s}_t$\footnote{$\hat{.}$ is used to distinguish the state prediction $\hat{s}_t$ from the true state of the environment $s_t$ which is assumed not directly observable by the agent.} is the representation of the true state $s_t$ at time step $t$.

Learning a good state representation in unsupervised manners is a non-trivial challenge, as the state representation should contain all, and only, the information which is relevant for allowing the agent to improve the policy and to solve the task. At the same time, the dimensionality of the state should be kept to a minimum to improve the agent's training efficiency. The state representation should also map observations to states in unambiguous ways, i.e. the state representation should be Markovian \cite{markovianSRL}. Eventually, the observation-state mapping should generalise to unseen observations with similar properties to the seen ones.

Many methods for unsupervised or self-supervised learning of state representations using neural networks have been developed over the years. However, three major categories of approaches can be distinguished accordingly to \cite{srlsurvey}. 

The first category includes all the methods of encoding information to low-dimensional spaces by relying only on observation reconstruction. In this context,  Auto-Encoders, or AEs, variational AEs, or denoising AEs \cite{autoencoders}, \cite{VAE_1} and \cite{VAE_2_DAE} can be used for learning to reconstruct the input observations that are fed through a neural network, composed of an encoder $\phi$ and a decoder $\phi^{-1}$, with a bottleneck. The bottleneck is used as the state vector for learning the policies. Despite the success, the main downside of these approaches is that AEs usually do not distinguish between task-relevant information and task-irrelevant information, such as background textures in images. This framework is known for ignoring small objects present in the observations, while these objects can be relevant for solving the task. Moreover, the observation reconstructions are not usually used by the reinforcement learning algorithms making the decoder unnecessary complexity \cite{PlannableMDPhomo}.

The second category of approaches for learning state representations takes advantage of the learning of the MDP model, i.e. the forward transition model, the reward model, and the inverse model. These approaches often use only an encoder that is trained to predict the next state embedding given the current state, the reward, and the action taken.
Because no decoder is used, the representation may collapse to trivial solutions, especially in case of sparse rewards \cite{PlannableMDPhomo}, \cite{DeepMDP}. Therefore, often these dynamical models are combined with the auto-encoder framework in order to improve and prevent collapsing of the learned state representation or with contrastive losses \cite{constrastiveloss}. 

Eventually, the third category includes all the methods loosely constraining the state space using auxiliary loss functions injecting prior knowledge in the form of loss functions for training the encoder networks. The so-called robotics priors have been introduced in \cite{Jonschkowski2015} as loss functions for encoding prior knowledge of the world, e.g. the physical laws, into the learning of the state representation for different robotics navigation tasks. Priors-based methods allow high sample efficiency and robustness of the learned representation and consequently of the policies. In this work, we focus on this category of approaches.

%% file: related_work.tex
The concept of robotics priors is introduced by \cite{Jonschkowski2015} for solving the problem of unsupervised learning of an informative state representation in the context of simple robotics navigation tasks. In those experiments, the agent's action space is chosen to be discrete, and the mobile robot can only move forward, backward, left or right at each time step. Furthermore, the robot is equipped with an RGB camera with a field of a view of $\SI{300}{\degree}$.

In \cite{botteghi2020}, the priors are adapted to incorporate the reward properties better, improve the quality of the learned state representation and, consequently, the quality of the learned reinforcement learning policy. 

In \cite{Jonschkowski2017}, the authors propose the so-called position-velocity encoder for learning a valid state representation. This work introduced new prior losses that exploit the relation between the position and the velocity of the inverted pendulum, the cart-pole and the ball-in-a-cup problems. Using the position-velocity encoder, all these different tasks can be efficiently learned from pixel inputs. 

In \cite{ReferencePointPrior}, the original set of robotics priors \cite{Jonschkowski2015} is evaluated in the presence of distractors, i.e. disturbing visual elements such as shadows or randomly moving backgrounds. When the domain randomization is strong, the robotic priors struggle to construct a coherent state representation. Therefore, the authors propose the reference-point prior, using true state values, to regularize the learned state space and mitigate such a limitation. 

Eventually, in  \cite{LandmarkPrior2019}, the authors propose an approach for learning state representation using the robotics priors in the case of more complex environments where a single observation is not enough to distinguish between two or more states. Therefore, they employ a recurrent LSTM-based encoder, mapping sequences of observations to single state predictions. Moreover, to obtain a coherent representation, they extend the reference-points prior, proposed in \cite{ReferencePointPrior}, to include multiple reference points, or landmarks, to connect the state predictions from different trajectories. 

Differently from all these approaches, we study the problem of learning state representations in the context of continuous action spaces and by exploiting the underlying action structure when constructing the priors, i.e. the loss functions. Here, we do not consider heavy domain randomization nor the problem of a recurrent state representation. However, we focus on a purely unsupervised approach that does not require the knowledge of true state values. Our approach can be considered orthogonal to \cite{ReferencePointPrior} and \cite{LandmarkPrior2019}. The proposed priors can be directly combined with the reference-points prior and with a recurrent encoder network. Moreover, differently from \cite{Jonschkowski2017}, \cite{ReferencePointPrior}, \cite{LandmarkPrior2019}, we test our approach on a real robot.



Eventually, all these approaches study the problem of state representation learning by introducing different robotics priors with the underlying assumption of discrete action spaces. In our work, we study the problem of state representation learning in continuous action spaces. Moreover, we aim at exploiting the structure of the action space to improve the quality of the learned state representation and, consequently, the efficiency of the policy learning.

%% file: methodology.tex
\subsection{Proposed Approach}

\begin{figure*}[!ht]
\centering
\begin{subfigure}{0.19\textwidth}
  \centering
  \includegraphics[width=0.60\textwidth]{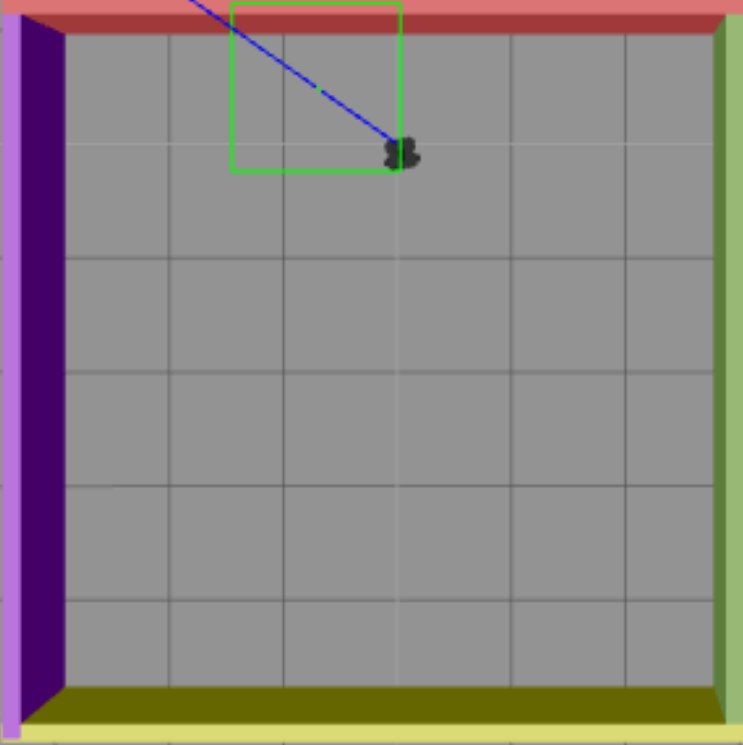}
  \captionsetup{justification=centering}
  \caption{\textit{Env-1}}
  \label{fig:env1}
\end{subfigure}
\begin{subfigure}{0.19\textwidth}
  \centering
  \includegraphics[width=0.63\textwidth]{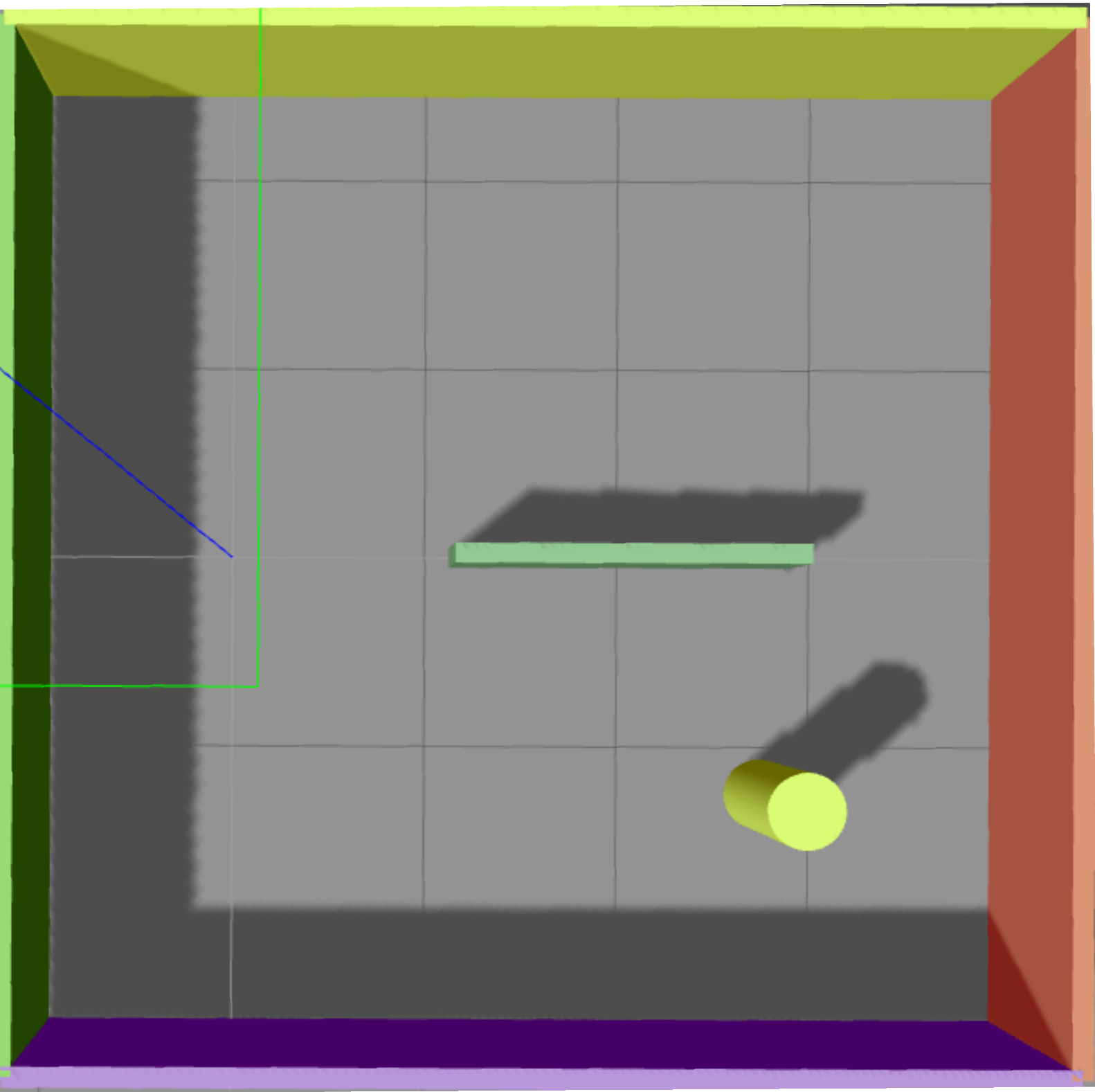}
  \captionsetup{justification=centering}
  \caption{\textit{Env-2}}
  \label{fig:env2}
  \end{subfigure}
 \begin{subfigure}{0.19\textwidth}
  \centering
  \includegraphics[width=0.7\linewidth]{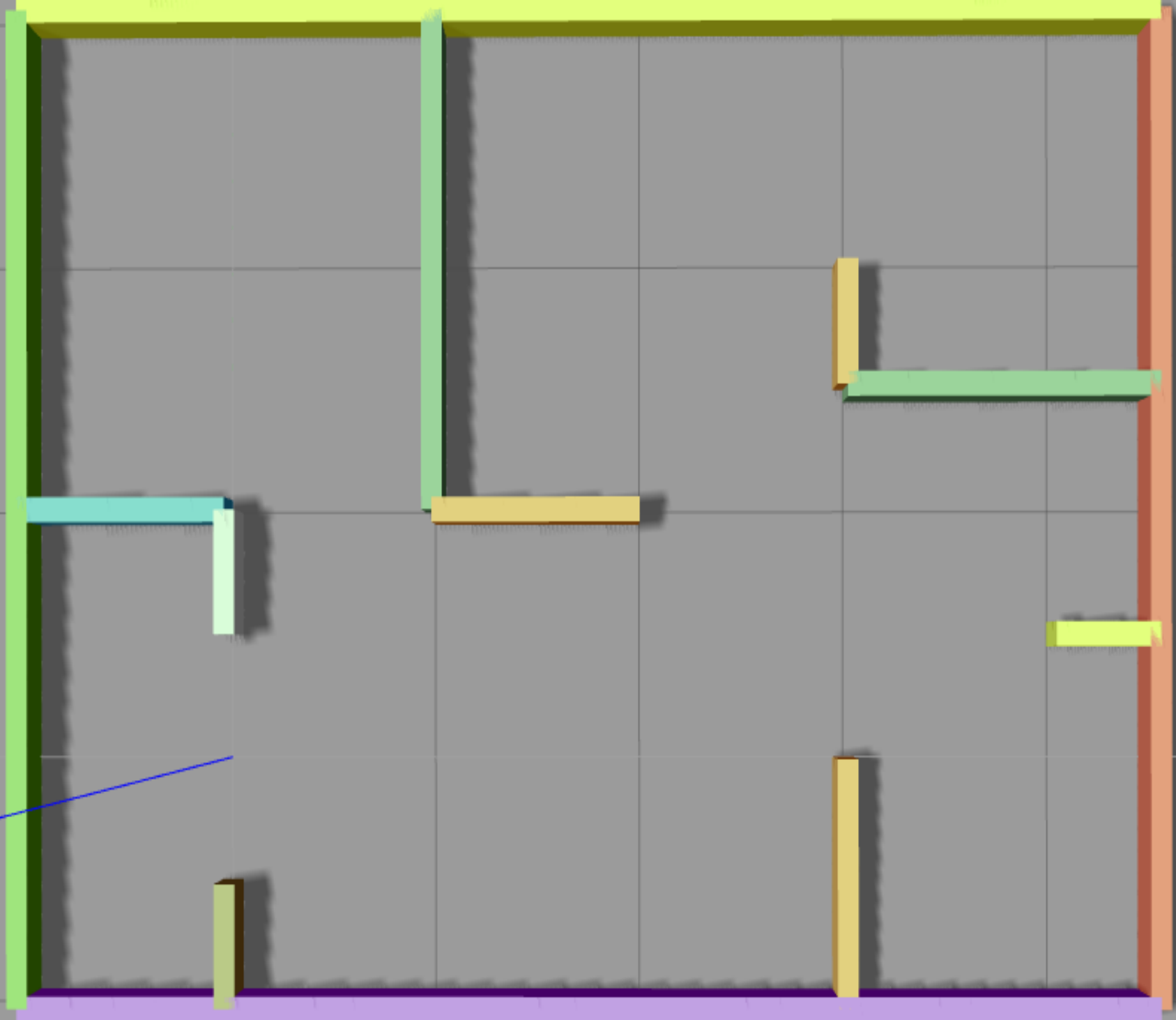}
  \captionsetup{justification=centering}
  \caption{\textit{Env-3}}
  \label{fig:env3}
  \end{subfigure}
 \begin{subfigure}{0.19\textwidth}
  \centering
  \includegraphics[width=0.6\linewidth]{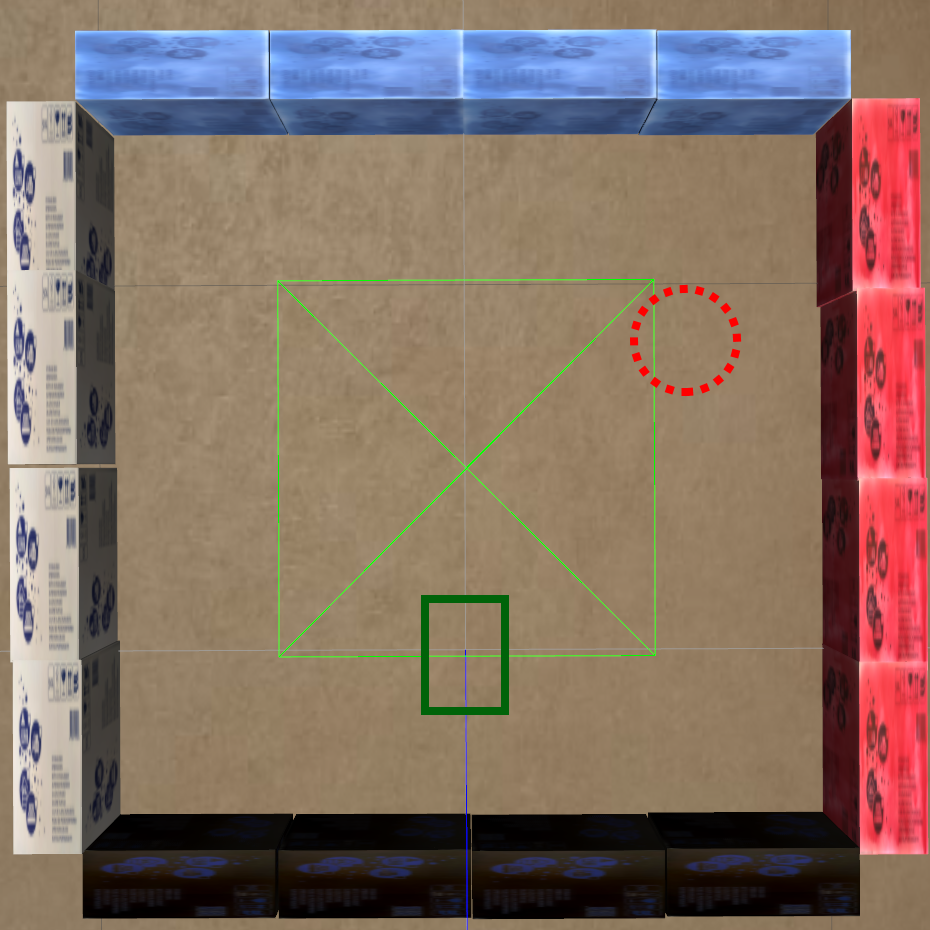}
  \captionsetup{justification=centering}
  \caption{\textit{Env-4}}
  \label{fig:env4}
  \end{subfigure}
  \begin{subfigure}{0.19\textwidth}
  \centering
  \includegraphics[width=0.6\linewidth]{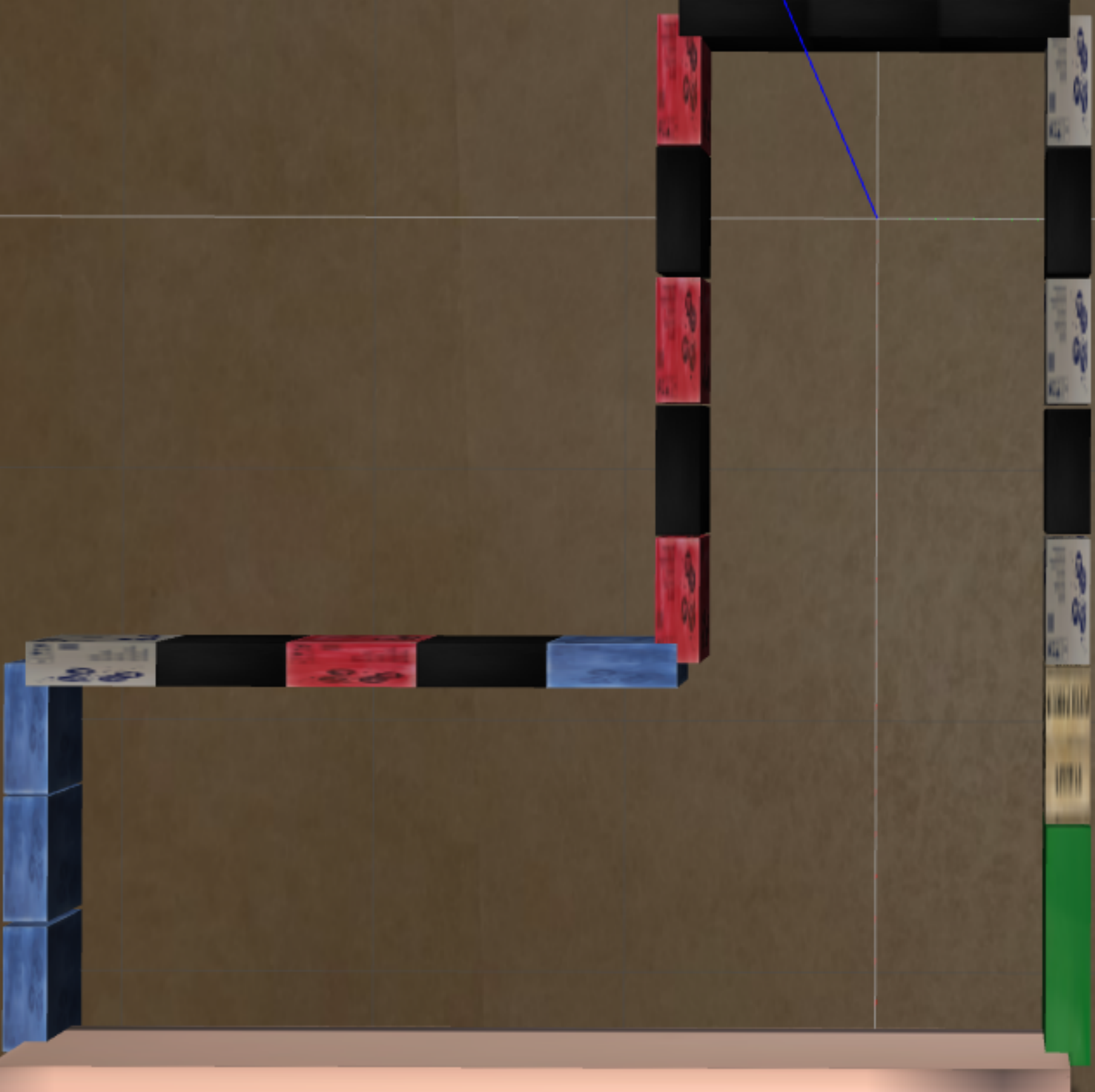}
  \captionsetup{justification=centering}
  \caption{\textit{Env-5}}
  \label{fig:env5}
  \end{subfigure}
\captionsetup{justification=centering}
\caption{Examples of simulation environments.} 
\label{fig:sim_envs}
\end{figure*}


We propose a framework for sample-efficient end-to-end reinforcement learning for solving different robotics navigation tasks directly from high-dimensional sensory readings. The framework is composed of two building blocks: first, a low-dimensional state representation is learned from the observations, and then the optimal policy is learned solely based on such state representation, as shown in Figure \ref{fig:state_representation_framework}. 
The state representation is learned in an unsupervised fashion by introducing a new set of robotics priors. The robotics priors are a way to inject prior knowledge into the state representation learning step by loosely constrain the learned state space. Our work extends the concept of robotic priors to continuous action spaces by exploiting the underlying structure of the actions into the encoding step of the state representation learning. 

For learning the optimal policy, we choose deep deterministic policy gradient, DDPG, as the candidate RL algorithm for its ability to deal with continuous state and action spaces. However, it is worth mentioning that the proposed approach is independent of the specific RL algorithm and any algorithm that can deal with continuous action space can be in principle employed.

With this work, we aim at addressing the following research questions:
\begin{itemize}
    \item How can we take advantage of the underlying action structure of continuous action spaces for learning the state representation using the robotic priors?
    \item To what extent can the priors, exploiting actions structure, be beneficial for policy learning?
    \item To what extent can the learned state representations, using the proposed priors, and the policies learned from such representations be transferred to real robots without further re-training using real-world data?
\end{itemize}

\subsection{Robotics Priors for Continuous Action Spaces}

Unlike discrete action spaces, continuous actions spaces have an underlying structure that we can exploit when learning a low-dimensional state representation. In many situations, observation changes and, consequently, agent's state changes are directly related to the magnitude of the action taken. This simple concept can be exploited and used as a loose constraint when learning the state representation using the robotics priors.   
The new set of robotics priors is presented below, where $\hat{s}_t$ corresponds to the state prediction given observation $o_t$, $a_t$ is the action and $\Delta \hat{s}_t = \hat{s}_{t+1} - \hat{s}_t$.

\textit{\textbf{Simplicity Prior}}: The task-relevant information lies in a space with dimensionality much smaller than the sensory observations.

\textit{\textbf{Temporal coherence prior}}: 
The temporal coherence prior, introduced in \cite{Jonschkowski2015}, encodes the property that state changes are slow and states close in time should also be close in space. However, this prior treats all the state pairs similarly and does not consider the magnitude of the action taken. In our approach, we use the magnitude of the action $a_t$, connecting state prediction $\hat{s}_t$ and next state prediction $\hat{s}_{t+1}$, as weighting factor for the state distance $|| \Delta \hat{s}_t ||$. In particular, when an action with a large magnitude connects two state predictions, the loss function does not enforce the states to be as close as when an action with a small magnitude is taken. In this way, we can better exploit the structure of the smooth and continuous action space. This intuition leads to the prior in Equation (\ref{eq:Prior_TempCorr_cont2}).
\begin{equation}
\mathcal{L}_{\text{temp}}=\mathbb{E}\left[ \big(\| \Delta \hat{s}_{t} \| e^{-\alpha\|a_t\|} \big)^2 \right]   
\label{eq:Prior_TempCorr_cont2}
\end{equation}
where the hyper-parameter $\alpha$ is used to weight the effect of the action magnitude on the state difference. 

\textit{\textbf{Proportionality prior}}: 
The original proportionality prior \cite{Jonschkowski2015} encodes the heuristic that the state variation of two different state pairs should be similar if the actions taken are similar. With similar reasoning to the case of the temporal coherence prior, in continuous action spaces, the similarity in states property translates into an additional weighting factor dependent on the difference in magnitude between the actions  $a_{t_1}$ and $a_{t_2}$. The action difference scales the need of minimizing the state difference. The new proportionality prior is shown in Equation (\ref{eq:Prior_proportial_cont2}).
\begin{equation}
\mathcal{L}_{\text{prop}}=\mathbb{E}\left[ \big(\| \Delta \hat{s}_{t_2} \| - \| \Delta \hat{s}_{t_1} \|\big)^2 e^{-\beta \|a_{t_1}-a_{t_2}\|^2}  \right]   
\label{eq:Prior_proportial_cont2}
\end{equation}
where hyper-parameter $\beta$ is used to weight the effect of the action difference on the state difference. 

\textit{\textbf{Repeatability prior}}: The repeatability prior reinforces the similarity of states not only in magnitude but also in direction. In particular, if two different states are similar and similar actions are taken in each of them, the magnitude of difference in the transition $\Delta s_{t_1}$ and $\Delta s_{t_2}$ should be limited. The new repeatability is shown in Equation (\ref{eq:Prior_Repeatablity_cont2}).
\begin{equation}
\mathcal{L}_{\text{rep}}=\mathbb{E}\left[ \| \Delta \hat{s}_{t_2} - \Delta \hat{s}_{t_1} \|^2 e^{-||\hat{s}_{t_2} - \hat{s}_{t_1}||^2} e^{-\beta ||a_{t_1}-a_{t_2}||^2}\right] 
\label{eq:Prior_Repeatablity_cont2}
\end{equation}

\textit{\textbf{Causality prior}}:
The temporal coherence and proportionality are aggregating priors that tend to reduce the distance between the state predictions. Therefore, to prevent the trivial mapping in which all the states are mapped to the origin, we employ a contrastive loss pushing apart state predictions. We further enhance the so-called causality prior, with an additional penalty dependent on the action difference.  This prior is shown in Equation (\ref{eq:Prior_Causality_cont2}).
\begin{equation}
\mathcal{L}_{\text{caus}}=\mathbb{E}\left[ e^{-||\hat{s}_{t_2} - \hat{s}_{t_1}||^2} e^{-\beta ||a_{t_1}-a_{t_2}||^2}  \right]
\label{eq:Prior_Causality_cont2}
\end{equation}

\textit{\textbf{Total loss}}:
The overall loss function that is minimised for learning the state representation is shown in Equation (\ref{eq:total_loss}).
\begin{equation}
\mathcal{L} = \omega_1\mathcal{L}_{temp} + \omega_2\mathcal{L}_{prop} + \omega_3\mathcal{L}_{rep} + \omega_4\mathcal{L}_{caus} + \omega_5\mathcal{L}_{reg} 
\label{eq:total_loss}
\end{equation}
where $\omega_1 = 1$, $\omega_2 = 1$, $\omega_3 = 1$, $\omega_4 = 2$, $\omega_5 = 1$ and $L_{reg}$ corresponds to the $L_2$ regularization loss. The weights of the single loss functions are chosen by performing a grid search $\in \{1, 2, 3, 4, 5\}$. However, the method is not very sensible to the choice of the hyperparameters and other sets of weights may produce similar results.

\subsection{Neural Network Architectures}

\begin{figure*}[h!]
    \centering
    \begin{subfigure}[0.33]{0.65\columnwidth}
    \includegraphics[width=\linewidth]{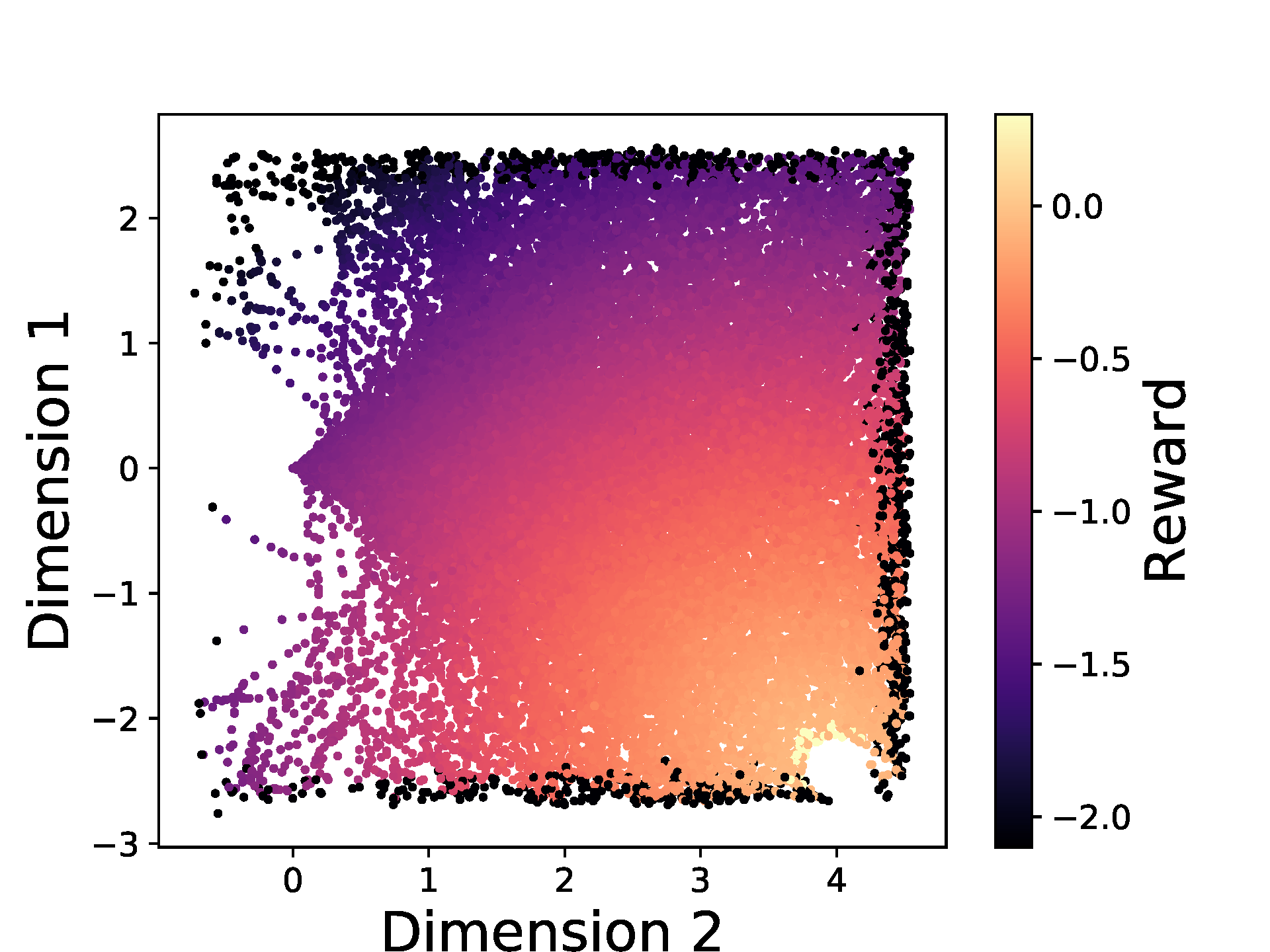}
    \caption{True state values}
    \label{fig:0}
  \end{subfigure}
  \hfill
\begin{subfigure}[0.33]{0.65\columnwidth}
    \includegraphics[width=\linewidth]{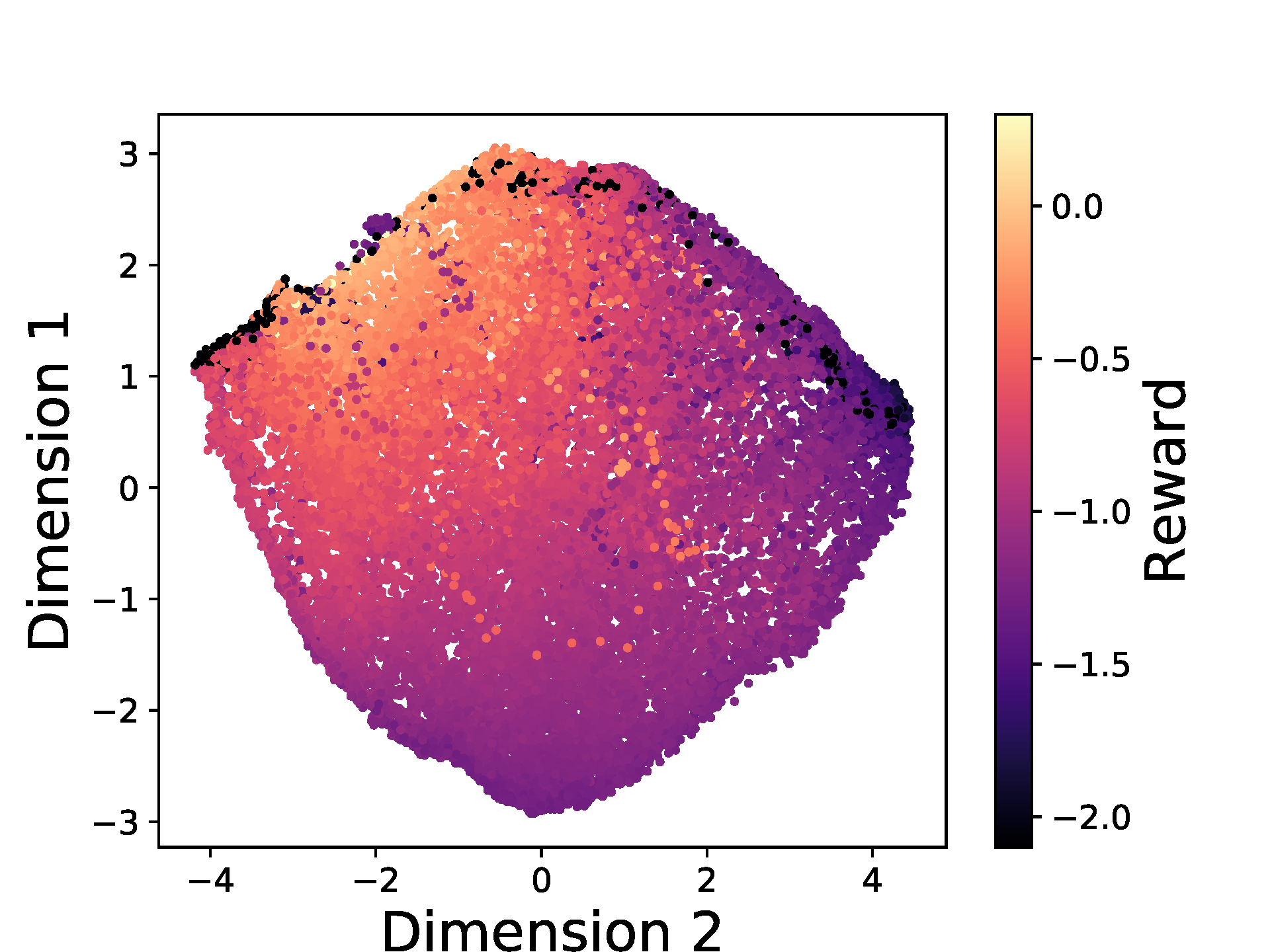}
    \caption{Our Priors}
    \label{fig:1}
  \end{subfigure}
  \hfill 
  \begin{subfigure}[0.33]{0.65\columnwidth}
    \includegraphics[width=\linewidth]{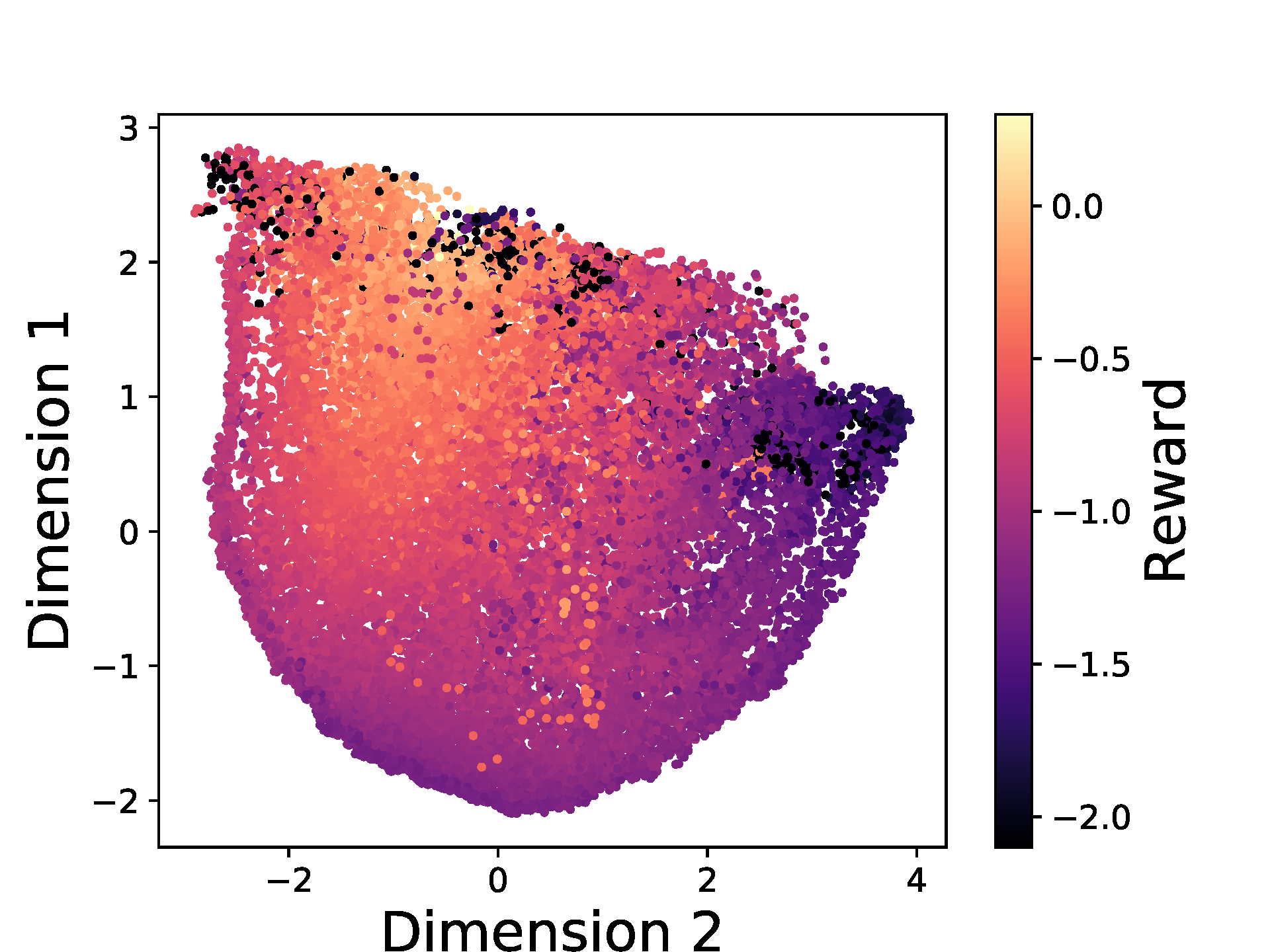}
    \caption{Priors in \cite{botteghi2020}}
    \label{fig:2}
  \end{subfigure}
   \hfill 
  \begin{subfigure}[0.33]{0.65\columnwidth}
    \includegraphics[width=\linewidth]{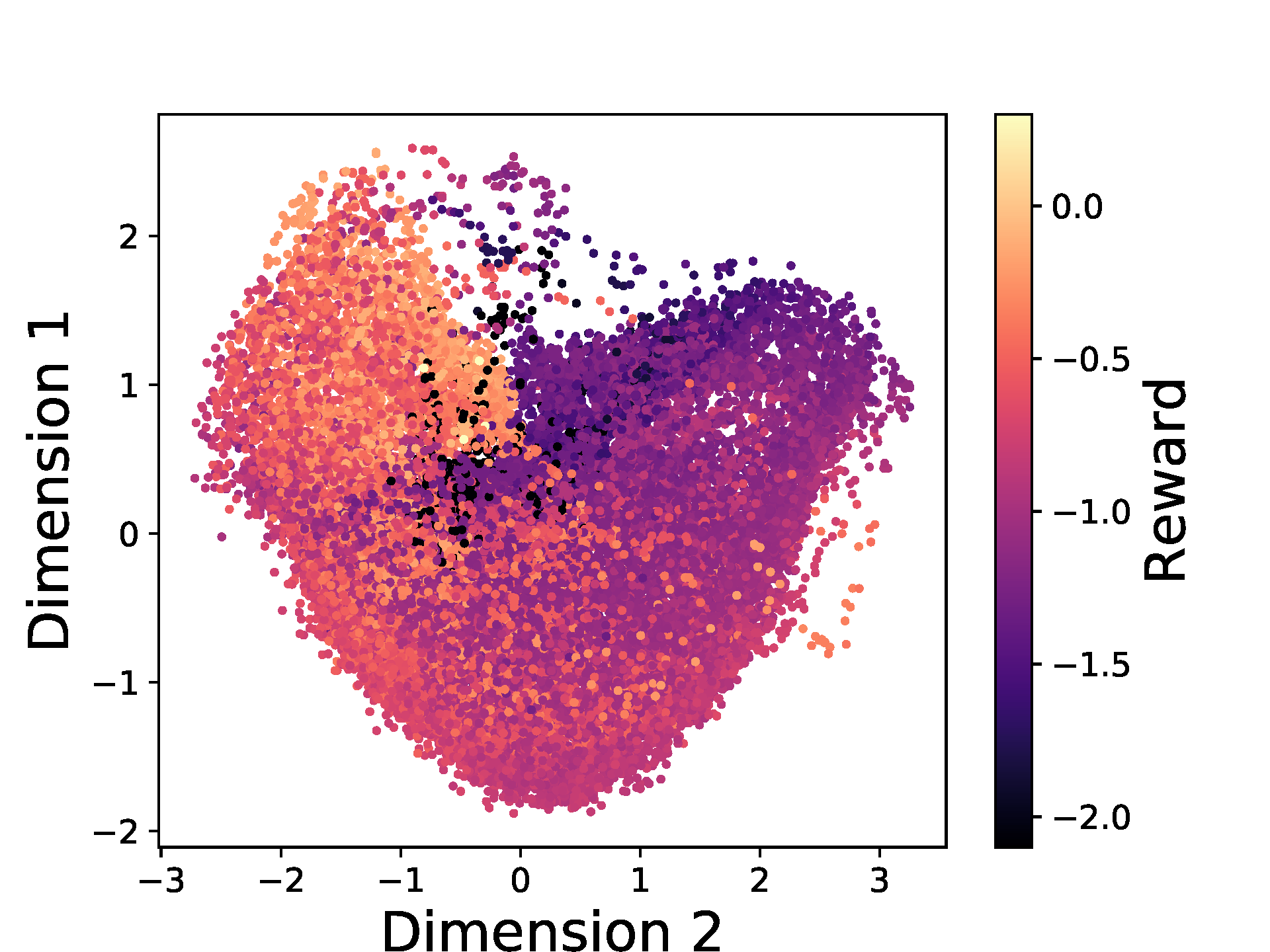}
    \caption{Adaptation of the priors in \cite{Jonschkowski2015}}
    \label{fig:3}
  \end{subfigure}
     \hfill 
  \begin{subfigure}[0.33]{0.65\columnwidth}
    \includegraphics[width=\linewidth]{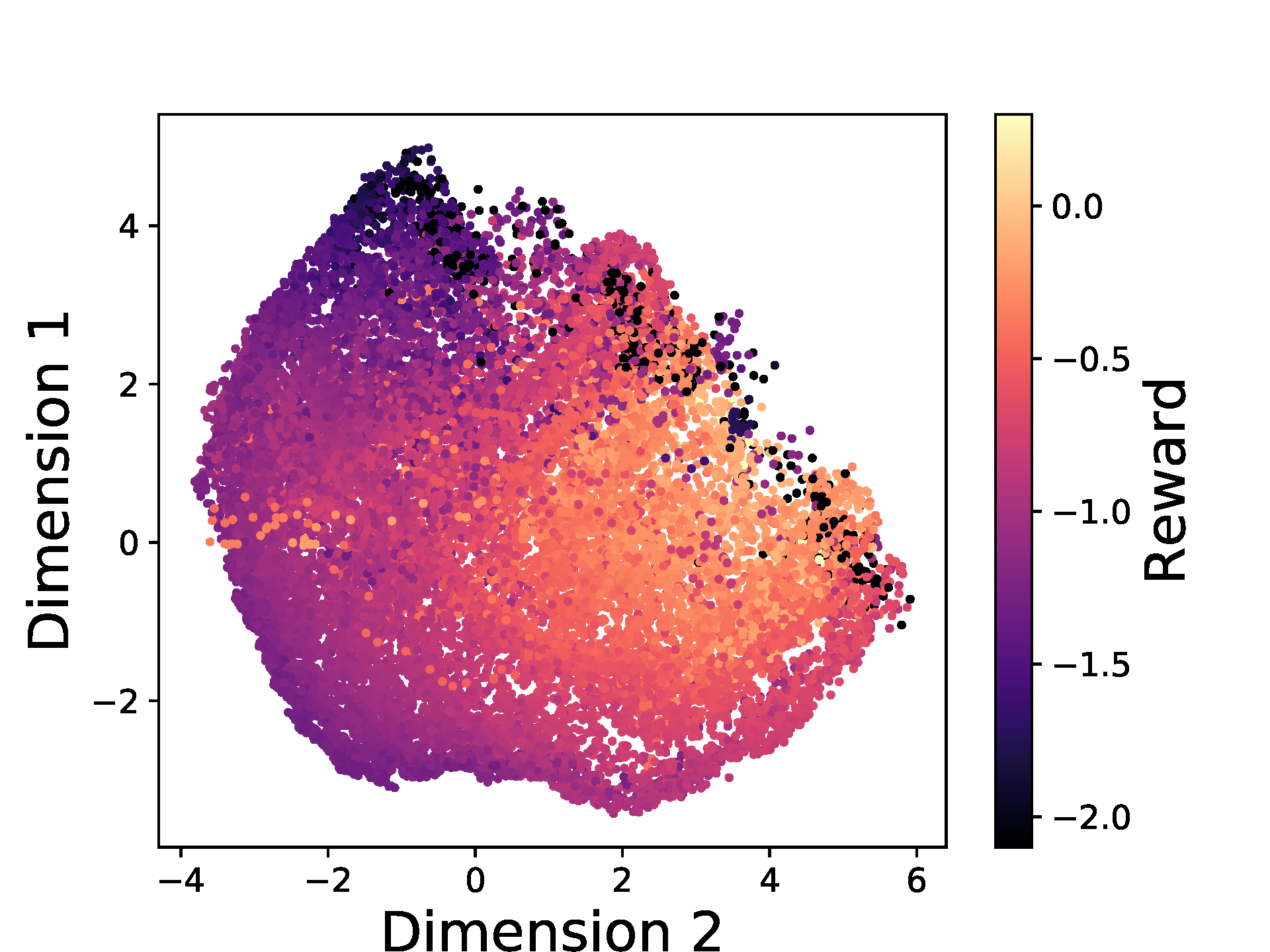}
    \caption{Adaptation of the priors in \cite{LandmarkPrior2019}}
    \label{fig:4}
  \end{subfigure}
     \hfill 
  \begin{subfigure}[0.33]{0.65\columnwidth}
    \includegraphics[width=\linewidth]{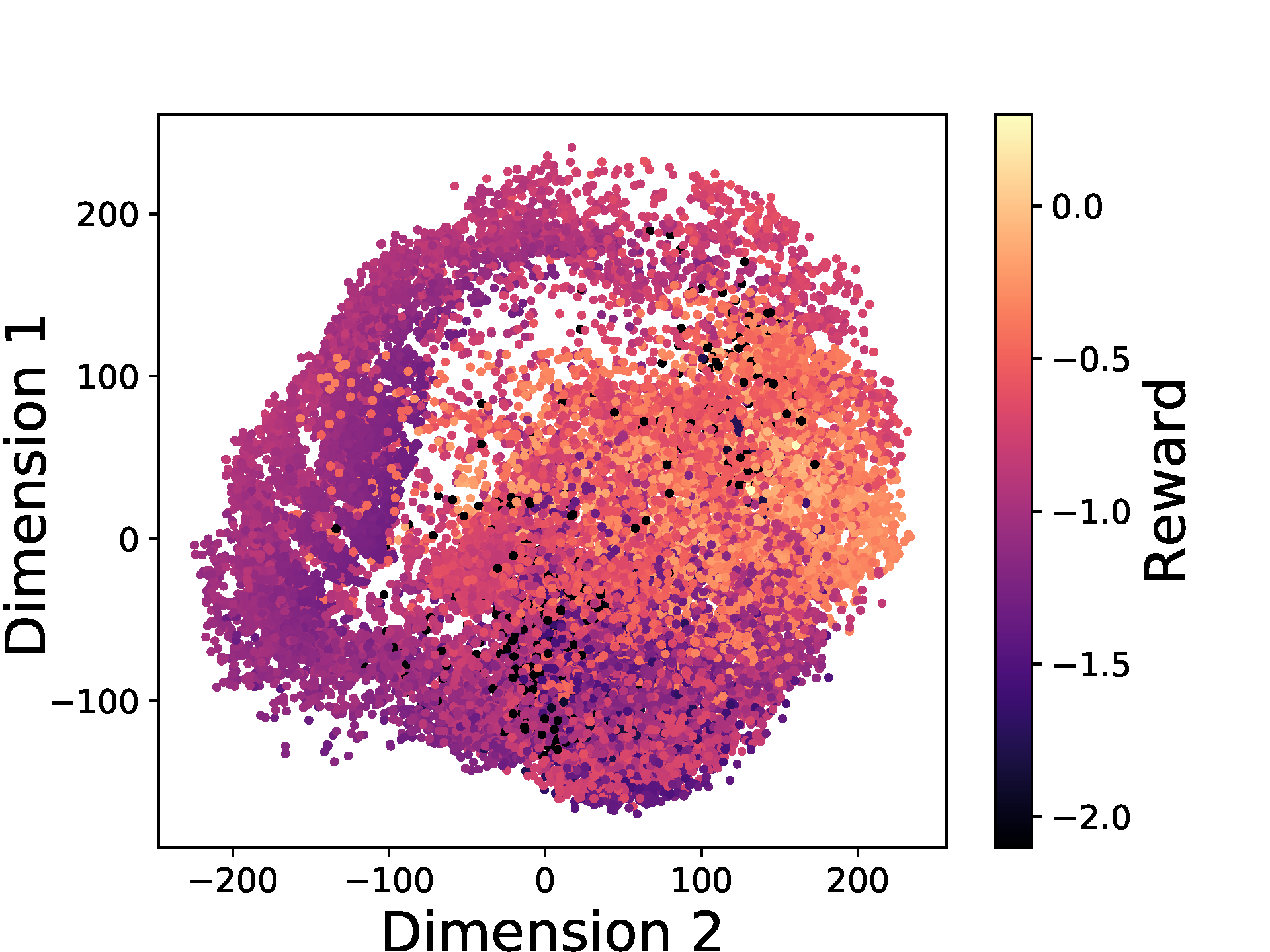}
    \caption{Auto-Encoder}
    \label{fig:5}
  \end{subfigure}
    \caption{True state values (Figure \ref{fig:0}), and first two principal components (Figure \ref{fig:1}-\ref{fig:5}), obtained with PCA, of learned state representations for environment \textit{Env-1} and target location in the bottom left corner (see Figure \ref{fig:env1}).}
    \label{fig:comparison_staterepresentation}
\end{figure*}

 

As shown in Figure \ref{fig:state_representation_framework}, the framework includes two mappings: from observations to low-dimensional state predictions and from state predictions to actions. Both mappings are learned through neural networks. In particular, the \textit{SRL network}
is responsible for the learning of the low-dimensional state representation from the multi-modal sensory observations. The architecture is similar to the one used in  \cite{botteghi2020}, where the sensor modalities are independently processed by convolutional layers, flattened and merged through fully connected layers to create the final low-dimensional state prediction of dimension 5. The state dimension is chosen accordingly to the studies done in \cite{botteghi2020} and \cite{LandmarkPrior2019}.
On the other side, the \textit{RL networks}, composed of an actor mapping states to actions, and a critic estimating the action-value function, are responsible for the learning of the optimal policy. Both architectures present three fully connected hidden layers of dimension 512. The output layer of the actor generates the linear and angular velocities set-point for the low-level controllers of the robot. We allow only forward motion by limiting the linear velocity with sigmoid activation. The output of the critic estimates the Q-value of the input state-action pair.



%% file: experimental_design.tex
\subsection{Mobile Robot Navigation with Multiple Sensor Modalities in Different Environments}

We evaluate the proposed approach by studying the problem of learning to control a differential-drive mobile robot, equipped with an RGB camera and a 2D LiDAR, to reach various target locations without collision with obstacles in different environments. Examples of environments are shown in Figure \ref{fig:sim_envs}. 

Since investigating the applicability of transfer learning is one of the goals of this work,
we first experiment in virtual simulated environments using the ROS-Gazebo platform, where we aim at learning the state representation and the optimal policy. Then, we transfer the learned models to the real robot without further training using real data. For the transfer learning experiments, we use the environments depicted in Figure \ref{fig:env4} and \ref{fig:env5}. The robot used is the TurtleBot 3 Waffle Pi. 

For all the experiments, we use a distance-based reward function, as shown in Equation (\ref{distance_rew}).
\begin{equation}
    \small
    \text{R}(s_t,a_t)=\begin{cases}
    r_{\text{reached}}, & d \leq d_{\text{min}},\\
    r_{\text{crashed}}, & s_t = s_{ts},  \\
    -\zeta (d_t - d_{t-1}), & \text{otherwise}.
    \end{cases}
    \label{distance_rew}
\end{equation}
where $d_t = \norm{p_t^{x,y}-g}_2$ is the distance from the current position ${p}_t$ of the robot at time $t$ with respect to the inertial frame and the target's location $g$ expressed in the robot's coordinate frame, $d_{t-1}$ the distance at time $t-1$, $r_\text{reached}$ is a bonus for reaching the target, $d_{\text{min}}$ is the minimum distance threshold below which the navigation target is considered reached, $\zeta$ is a scaling factor, and $r_\text{crashed}$ is a penalty for reaching a terminal state $s_{ts}$, e.g. hitting an obstacle or reaching the maximum number of steps in a single training episode. A distance-based reward function is a common choice for solving navigation tasks.


\subsection{Baselines}\label{sec:baselines}

\begin{figure*}[!ht]
\centering
\begin{subfigure}{0.27\textwidth}
  \centering
  \includegraphics[width=1.0\textwidth]{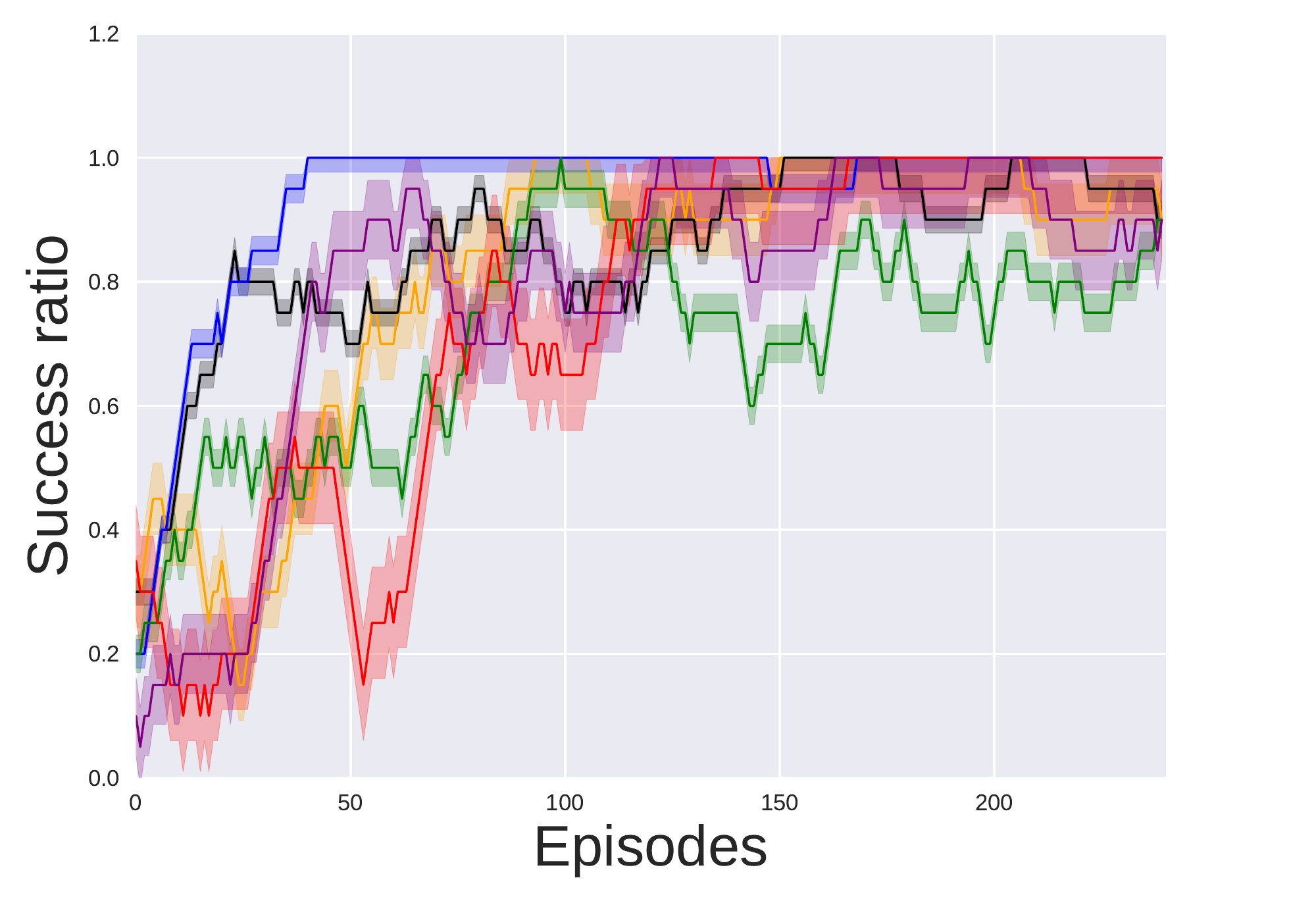}
  \captionsetup{justification=centering}
  \caption{\textit{Env-1}}
  \label{fig:res_env1}
\end{subfigure}
\hfill
\begin{subfigure}{0.27\textwidth}
  \centering
  \includegraphics[width=1.0\textwidth]{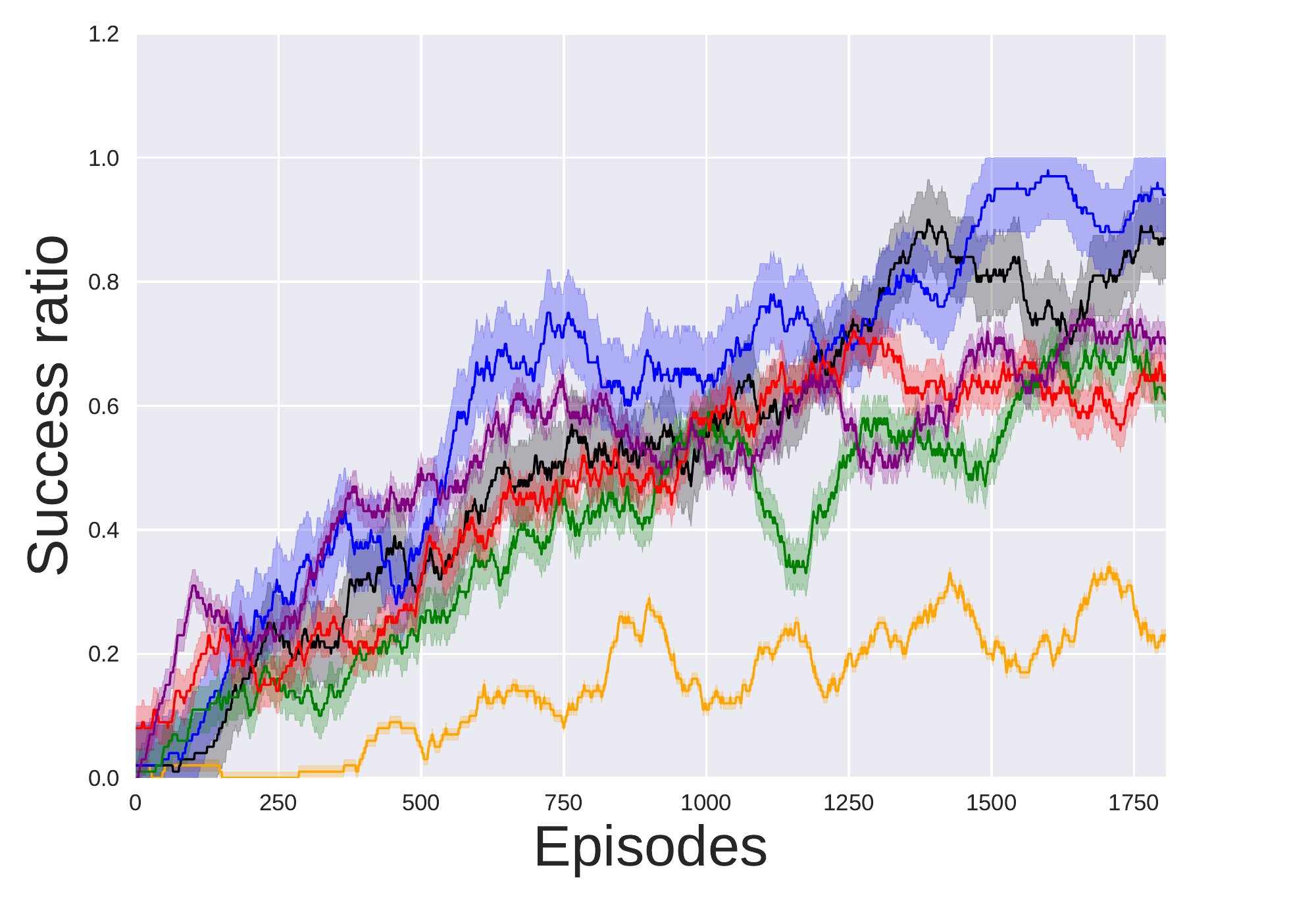}
  \captionsetup{justification=centering}
  \caption{\textit{Env-2}}
  \label{fig:res_env2}
  \end{subfigure}
 \hfill
 \begin{subfigure}{0.27\textwidth}
  \centering
  \includegraphics[width=1.0\linewidth]{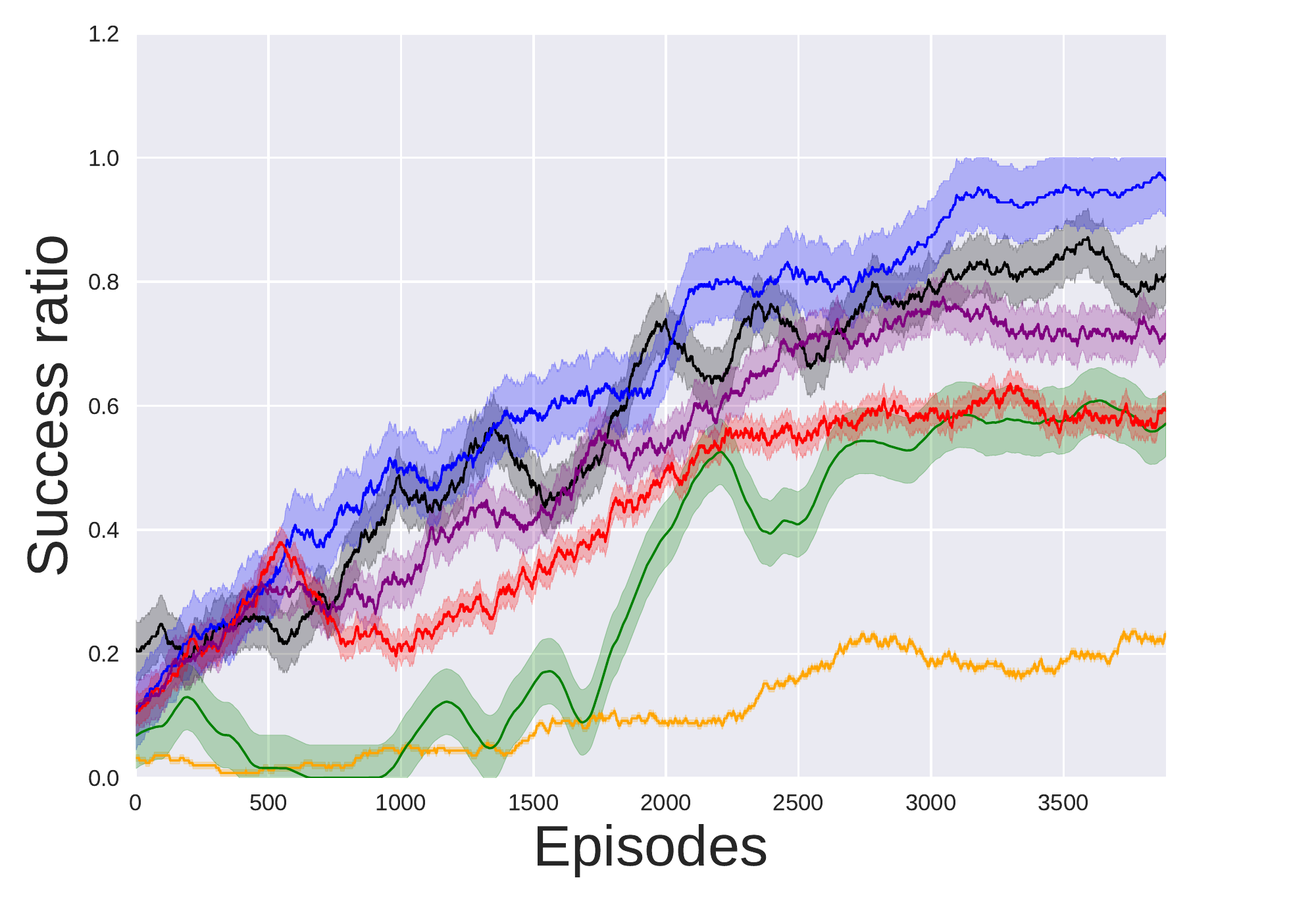}
  \captionsetup{justification=centering}
  \caption{\textit{Env-3} }
  \label{fig:res_env3}
\end{subfigure}
\hfill
\begin{subfigure}{0.15\textwidth}
  \centering
  \includegraphics[width=1.0\textwidth]{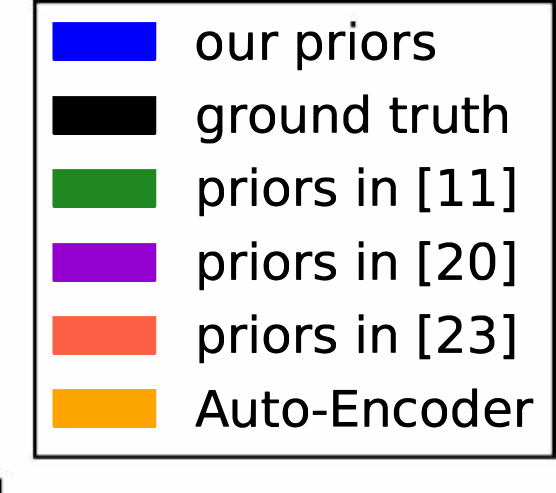}
  \captionsetup{justification=centering}
  \caption*{}
  \label{fig:res_legend}
\end{subfigure}
\hfill
\captionsetup{justification=centering}
\caption{Evolution of the success ratio over training in the different environments. The solid line represents the mean and the shaded area, the variance of the success ratio. For the sake of clarity, we omit the variance in Figure \ref{fig:res_env1}.} 
\label{fig:res_sim_envs}
\end{figure*}

We compare the proposed approach for learning the state representation with:
\begin{itemize}
    \item The adaptation of original robotics priors introduced in \cite{Jonschkowski2015} for continuous action spaces, where the action equality is replaced with a similarity between the action pairs, i.e. two actions are similar if the difference of their magnitudes is below a given threshold.
    \item The reward-shaped robotics priors proposed in \cite{botteghi2020}.
    \item The robotics priors proposed in  \cite{LandmarkPrior2019}, where, for fairness of the comparison, we remove the landmark prior that requires true state values.
    \item An Auto-Encoder (AE), where the latent code of the AE is used as an input to the reinforcement learning networks.
\end{itemize}
For each environment, we collect a data set comprised of randomly generated trajectories. We first learn a state representation of dimension 5 for each approach by updating the \textit{SRL network} for 20 epochs. In the AE case, the state dimension is set to 20, and the training lasts 200 epochs. Because our goal is to reach any possible target in a given environment, at each training episode, we randomly change the goal location and the starting pose of the robot to prevent getting biased by the environment structure. Therefore, to inform the agent of the target information, we concatenate the learned state predictions to the $(x,y)$-coordinates of the target position. Eventually, for further training stability, the previous action taken by the agent is also added to the complete extended state vector to improve the smoothness of the resulting trajectories. 

We analyse the quality of the learned state representations and the consequent performances in terms of success ratio\footnote{The success is defined by the robot reaching the target without collisions.} over the training of the reinforcement learning agent when fed with the learned representations. For fairness of comparison, the same \textit{SRL} and \textit{RL networks} architectures are used for each method, with the exception of the AE, where a decoder network is added to reconstruct the input data from the latent code.

Additionally, we compare with the agent trained using the true state, i.e. the pose of the robot. We concatenate the target position and the previous action taken for fairness of comparison to the true pose.

%% file: results.tex
\subsection{Analysis of the Learned State Representations}
In Figure \ref{fig:comparison_staterepresentation}, we show, through the plotting of the true states, i.e. the true robot's poses, and the first two principal components computed using PCA \cite{PCA}, of the different state representations obtained when training the \textit{SRL network} with the different baselines in environment \textit{Env-1}. 
Due to the additional regularization provided by the action terms in the priors (see Equation (\ref{eq:Prior_TempCorr_cont2})-(\ref{eq:Prior_Causality_cont2})), the learned state space appears to be the smoothest and most coherent with the true state space. This result means that our approach better incorporates the properties of the true state space. In particular, the priors proposed in \cite{botteghi2020}, in Figure \ref{fig:2},  suffer from a slight lack of smoothness near obstacles due to sudden changes in the rewards. The original priors, proposed in \cite{Jonschkowski2015}, in Figure \ref{fig:3}, by forcefully aggregating states in which similar actions are taken, disrupt the intrinsic smoothness of state and action spaces. The prior in \cite{LandmarkPrior2019}, in Figure \ref{fig:4}, can achieve a good representation, even though less smooth when compared to our priors. Eventually, as expected, the AE, in Figure \ref{fig:5}, learns the least interpretable representation. The latent states are only aggregated based on similarities of the observations, and there are no guarantees that two consecutive yet different observations are mapped close to each other.


\subsection{Simulation Results}

To assess the quality of the learned state representations in quantitative manners, we compare the performance of the different reinforcement learning agents when trained on such representations. Figure \ref{fig:res_sim_envs} shows the success ratio over training of the agents for the different simulation environments depicted in Figure \ref{fig:env1}-\ref{fig:env3}. The agent trained with the proposed method for state representation learning outperforms the other baseline due to the smoother and more coherent state representation that betters incorporates the properties of the true state space.

\begin{figure}[h!]
\centering
\begin{subfigure}{0.25\textwidth}
  \centering
  \includegraphics[width=0.85\textwidth]{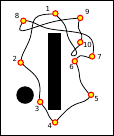}
  \captionsetup{justification=centering}
  \caption{\textit{Env-2}}
\end{subfigure}%
\begin{subfigure}{0.25\textwidth}
  \centering
  \includegraphics[width=0.85\linewidth]{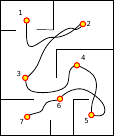}
  \captionsetup{justification=centering}
  \caption{\textit{Env-3}}
  \label{fig:sub2at2}
  \end{subfigure}%
  \captionsetup{justification=centering}
\caption{Trajectory tracking in the testing virtual environments. }
\label{traj}
  \end{figure}

Furthermore, in Figure \ref{traj}, we show the trajectories achieved after training by our approach. The robot has to reach a set of targets in \textit{Env-2} and \textit{Env-3} based on the sequence labelled in the Figure. The agent can successfully reach a sequence of different targets without collision with the obstacles.


    
    

\subsection{Real-World Experiments}

Eventually, we train the state representation and policy networks in the simulated environments \textit{Env-4} and \textit{Env-5}, depicted in Figure \ref{fig:env4} and \ref{fig:env5} respectively, that resemble the real-world environments. Then, we directly transfer them to the real robot without further retraining using real-world data. The learned low-dimensional state representation can robustly extract the most important features out of the sensory readings, and it can effectively reduce, if not cancel out the simulation-to-reality gap. Moreover, we test the robustness of the learned model against visual and depth distractors, such as lighting changes and a suddenly appearing moving object on the robot's path to the target. The complete video of our real-world experiments can be found at: \url{https://youtu.be/xujdA4b8-tY}.

%% file: conclusion.tex
This paper presents an end-to-end deep reinforcement learning framework that explicitly separates the learning of a low-dimensional state representation, given high-dimensional observations, with the policy learning for continuous state and action spaces. We show that we can exploit the underlying continuous action structure by means of the new robotics priors, in Equations (\ref{eq:Prior_TempCorr_cont2})-(\ref{eq:Prior_Causality_cont2}). The new priors allow the learning of a smoother and more coherent representation than the different robotics priors proposed in the literature. This translate into a higher success ratio of the learned policies when trained in different simulation environments. Eventually, the representations and policies learned in the simulation environment can be successfully transferred to the real world without any retraining using real-world data. The compression of high-dimensional observations into a low-dimensional state vector is the key element for transferring the learned models in the simulation environment to the real world.